\documentclass[fleqn,10pt]{wlscirep}
\usepackage[utf8]{inputenc}
\usepackage[T1]{fontenc}

\usepackage{subfigure}
\usepackage{multirow}
\usepackage{adjustbox}
\usepackage{pdfpages}
\usepackage{lastpage}
\usepackage{hyperref}

\title{Active shooter detection and robust tracking utilizing supplemental synthetic data}

\author[1]{Joshua R. Waite}
\author[2]{Jiale Feng}
\author[3]{Riley Tavassoli}
\author[3]{Laura Harris}
\author[1]{Sin Yong Tan}
\author[3]{Subhadeep Chakraborty}
\author[1,*]{Soumik Sarkar}
\affil[1]{Department of Mechanical Engineering, Iowa State University, Ames, IA 50011, USA}
\affil[2]{Department of Computer Science, Iowa State University, Ames, IA 50011, USA}
\affil[3]{Department of Mechanical Engineering, University of Tennessee, Knoxville, TN 37996, USA}

\affil[*]{soumiks@iastate.edu}

\begin{abstract}
The increasing concern surrounding gun violence in the United States has led to a focus on developing systems to improve public safety. One approach to developing such a system is to detect and track shooters, which would help prevent or mitigate the impact of violent incidents. In this paper, we proposed detecting shooters as a whole, rather than just guns, which would allow for improved tracking robustness, as obscuring the gun would no longer cause the system to lose sight of the threat. However, publicly available data on shooters is much more limited and challenging to create than a gun dataset alone. Therefore, we explore the use of domain randomization and transfer learning to improve the effectiveness of training with synthetic data obtained from Unreal Engine environments. This enables the model to be trained on a wider range of data, increasing its ability to generalize to different situations. Using these techniques with YOLOv8 and Deep OC-SORT, we implemented an initial version of a shooter tracking system capable of running on edge hardware, including both a Raspberry Pi and a Jetson Nano.
\end{abstract}
\begin{document}

\flushbottom
\maketitle
%
%
\thispagestyle{empty}


\section*{Introduction}
In 2005, the Federal Bureau of Investigation (FBI) and leading criminologists defined a ‘mass shooting’ as an attack in a public place where four or more victims were killed. Using this definition, there have been at least 149 public mass shootings across the United States since 1982~\cite{follman_aronsen_pan_2012}. While mass shootings are relatively rare, the impact they can have on a community, especially in the case of a school shooting, is more than enough reason to strive for improving public safety~\cite{ames}. However, the exact approach to improving school safety is heavily debated, often over concerns of invasion of privacy or degrading the quality of the student learning environment. On top of that, some of the common approaches, including metal detectors, armed school resource officers, and backpack searches, have shown varying effectiveness in different schools~\cite{doi:10.1080/15564886.2017.1307293}. A possible explanation for this is differences in implementations and resource availability. School resource officers can vary significantly from one school to another, sometimes serving a purely disciplinary role and other times serving a more supportive role~\cite{Madfis2016}.

Besides the common approaches mentioned above, there are also recent works that utilize advanced technology in detecting shooters.
One approach to shooter detection is using acoustic sensors in gunshot detection technology. \cite{valenzise2007scream, mares2021acoustic}. This type of system would simply alert law enforcement agencies when a gunshot is detected. However, this approach may have limitations in distinguishing between gunshots and other loud noises, such as fireworks or car backfires, and inaccurately capturing the exact location of a shooter moving or shooting from a distance 
Without visual information, this kind of shooter detection cannot provide details such as the number of shooters or their targets. Additionally, it requires the installation of additional specialized devices, which can be a disadvantage.


Our approach aims to minimize invasion of privacy as it would unobtrusively leverage video from existing security cameras. Additionally, our approach would automatically detect threats instead of relying on security personnel to monitor a surveillance system. As a result, the time taken to relay information about the shooter would be reduced. This can improve the effectiveness of law enforcement responding to the scene and be used to evacuate civilians when it is safe to move from cover. While schools are often the most discussed setting for this topic, our system could also be used in any public space where there is a risk of a shooting, such as hospitals, shopping malls, and airports.


While there is existing work and datasets focusing on the detection of guns or other weapons~\cite{abs-2105-01058,Narejo2021,app11136085,Ahmed2022DevelopmentAO,ashraf2022weapons,narejo2021weapon,9057329, MANIKANDAN2022104406, SALAZARGONZALEZ2020297, 10.1145/3154979.3154988}, the tracking of shooters, however, has not been extensively explored. Detecting the entire shooter has the potential to improve tracking robustness, as their location would not be as easily lost if the vision of the gun is obstructed. The challenge with this approach is that gathering good-quality data on shooters has proven time-consuming and difficult. Most publicly available data comprises poor-quality surveillance videos, often split into short, discontinuous clips unsuitable for training. Additionally, places where such a system would reasonably be implemented would likely already have good-quality security cameras. Thus, the data used to train should be of at least a reasonable baseline quality to avoid making the already difficult task of detecting guns more challenging than it needs to be. Another common type of video source used for this task is movies; however, the camera perspectives in movies are rarely similar to those that a security camera would capture. This is important to note because the appearance of a gun can change significantly, especially in the case of handguns, where they can appear to be nothing but a rectangle, similar to a smartphone. There are also privacy concerns with conducting experiments to record videos for the dataset, both for the individuals participating and the public buildings that would be used to simulate a shooting event.

As a result, we have explored the use of synthetic data generated with Unreal Engine to supplement the limited availability of real data. However, a well-known limitation of model training with synthetic data (even with the semi-realistic textures) is that the model does not directly transfer well to inference on real data. To improve the efficacy of training with synthetic data, we utilize domain randomization, which is a domain adaptation technique that aims to generalize a model through training with highly variable synthetic data~\cite{8202133,Tremblay_2018_CVPR_Workshops}. Besides that, transfer learning~\cite{9134370} also allows us to use various combinations of textured images of the Unreal Engine environment, masked images with random colors, and real data by sequentially training the detection model. We also augment the textured synthetic data with camera sensor effects to further help bridge the gap between synthetic and real data \cite{carlson2018modeling}. These effects include noise, blur, chromatic aberration, exposure, and color shift, which are randomly applied with varying strengths. 

\begin{figure}[h!]  
  \centering
  \includegraphics[width=1\linewidth]{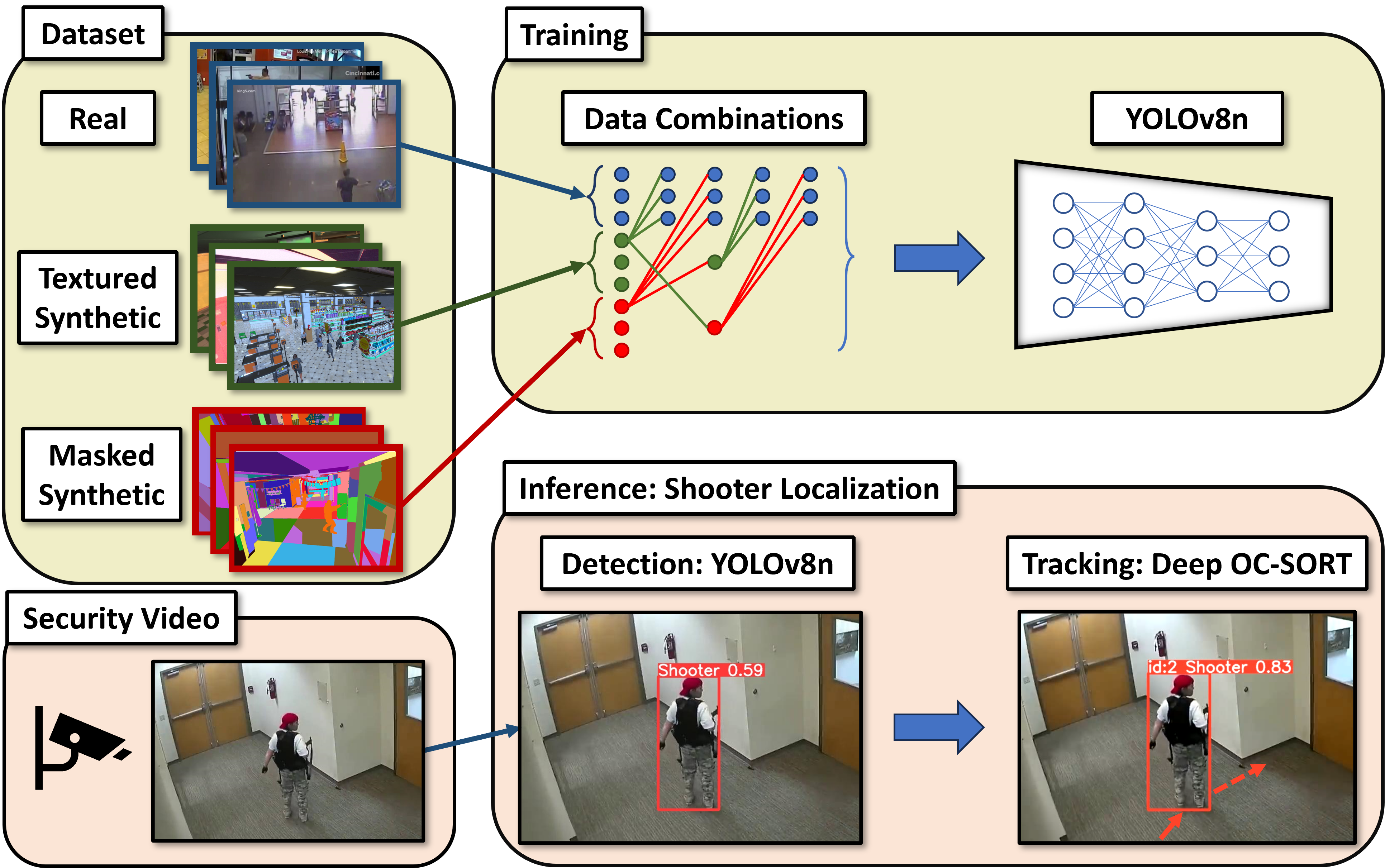}
  \caption{\label{fig:overview}
  An overview of the entire system. We train YOLOv8n using a combination of synthetic and real data. The best model is used for inference with Deep OC-SORT tracking to localize a shooter, enabling a faster and more informed response for law enforcement.}
\end{figure}

An overview of our system can be seen in Fig.~\ref{fig:overview}. It first shows the three types of data, real, textured synthetic, and masked synthetic, that comprise our shooter dataset. Various amounts of each type of data, the specifics of which are discussed later, are used sequentially to train YOLOv8n. The best performing YOLOv8n model is used with Deep Observation-Centric (OC)-SORT to track shooters in sources such as security videos, which can then be used to enable more informed law enforcement responses.

In terms of deployment, the use of edge devices, or small, relatively inexpensive computers, can also increase privacy by only transmitting necessary information, such as whether a threat is detected or not (binary), rather than the entire camera frames/images. On top of that, this also decreases the network bandwidth used for the system. Additionally, decentralized systems like this are generally more scalable and robust. While passing all video frames to a central server to be processed would be functionally the same, it would be more complicated to expand in the future, and the failure of the server would bring the whole system down. For example, if a building wanted to add additional cameras after the initial installation, they may be required to upgrade their entire server to handle the increased computational demand. On the other hand, since the edge devices can handle the computations in this system, they would only need additional devices to pair with the new cameras, providing a more straightforward cost estimation for expanding the system.



For the task of detecting and tracking shooters in public places, we make the following contributions:
\begin{itemize}
\item The creation of a publicly available dataset (synthetic and real) with annotations for gun and shooter classes will allow for further exploration of the detection and tracking of shooters.
\item Development of a robust tracking system utilizing gun detection-based shooter confirmation to reduce false positives while being more likely to keep track of a threat through occlusions.
\item Evaluation on implementing the proposed system on edge hardware, such as a Jetson Nano, and the considerations required.
\end{itemize}



\section*{Results}

Our results are broken into four primary subsections. We first present detection performance for when You Only Look Once v8 nano (YOLOv8n) models are trained with varying combinations of real, textured synthetic, and masked synthetic data. Next, we present the tracking performance using Deep OC-SORT with Omni-Scale Network (OSNET) Re-Identification (ReID) with and without gun confirmation for shooter IDs. We also analyze the system-level performance in a more realistic context rather than just using standard metrics. Lastly, we report the system's performance on edge devices such as a Jetson Nano and Rasberry Pi 4.

\subsection*{Detection with YOLOv8n}


\begin{figure}[ht!]
    \centering
    \subfigure[UE5]{\includegraphics[trim=0cm 0.4cm 0.5cm 0cm, clip, width=0.47\linewidth]{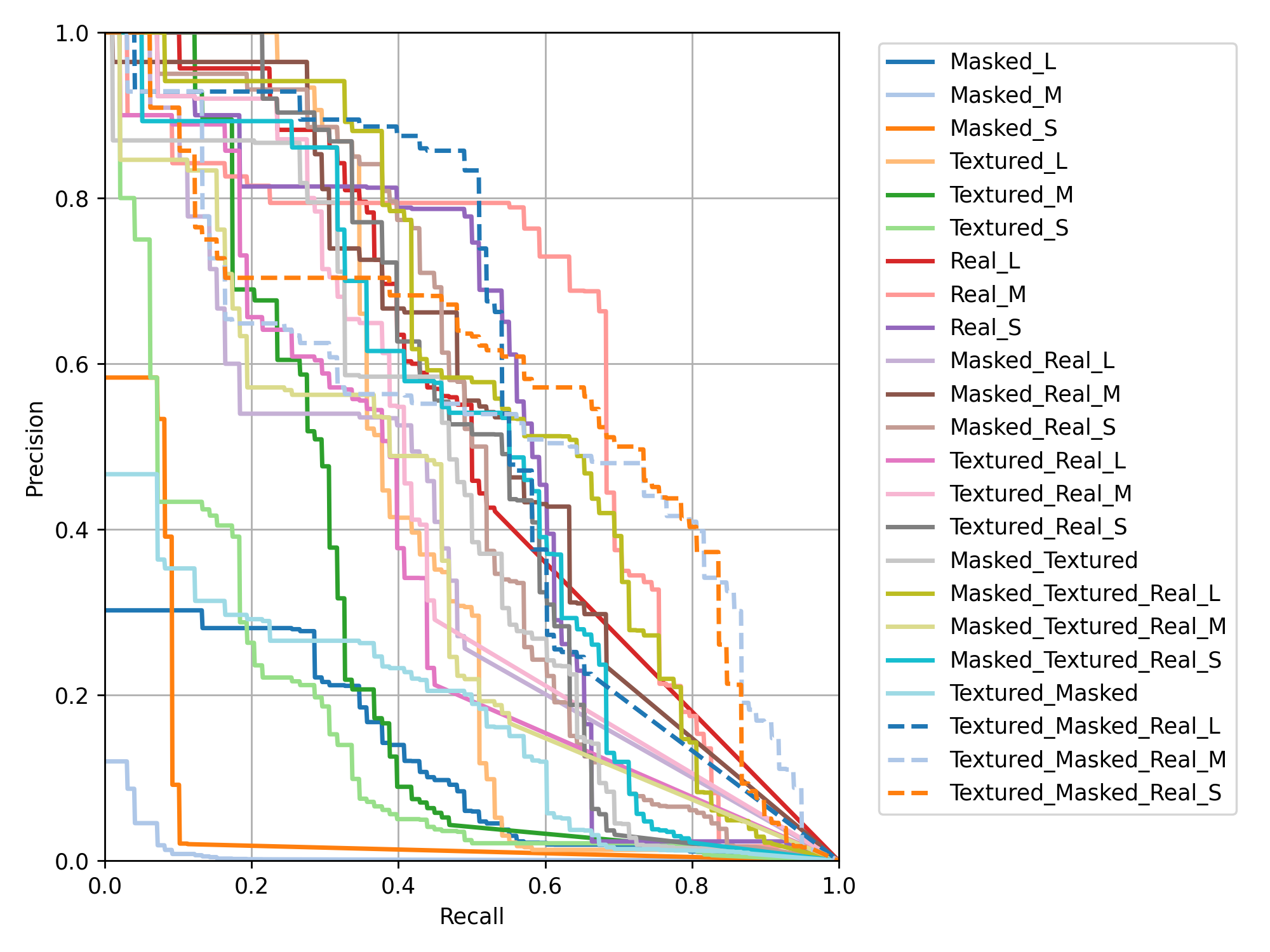}} 
    \subfigure[UE4 \& UE5]{\includegraphics[trim=0cm 0.4cm 0.5cm 0cm, clip, width=0.47\linewidth]{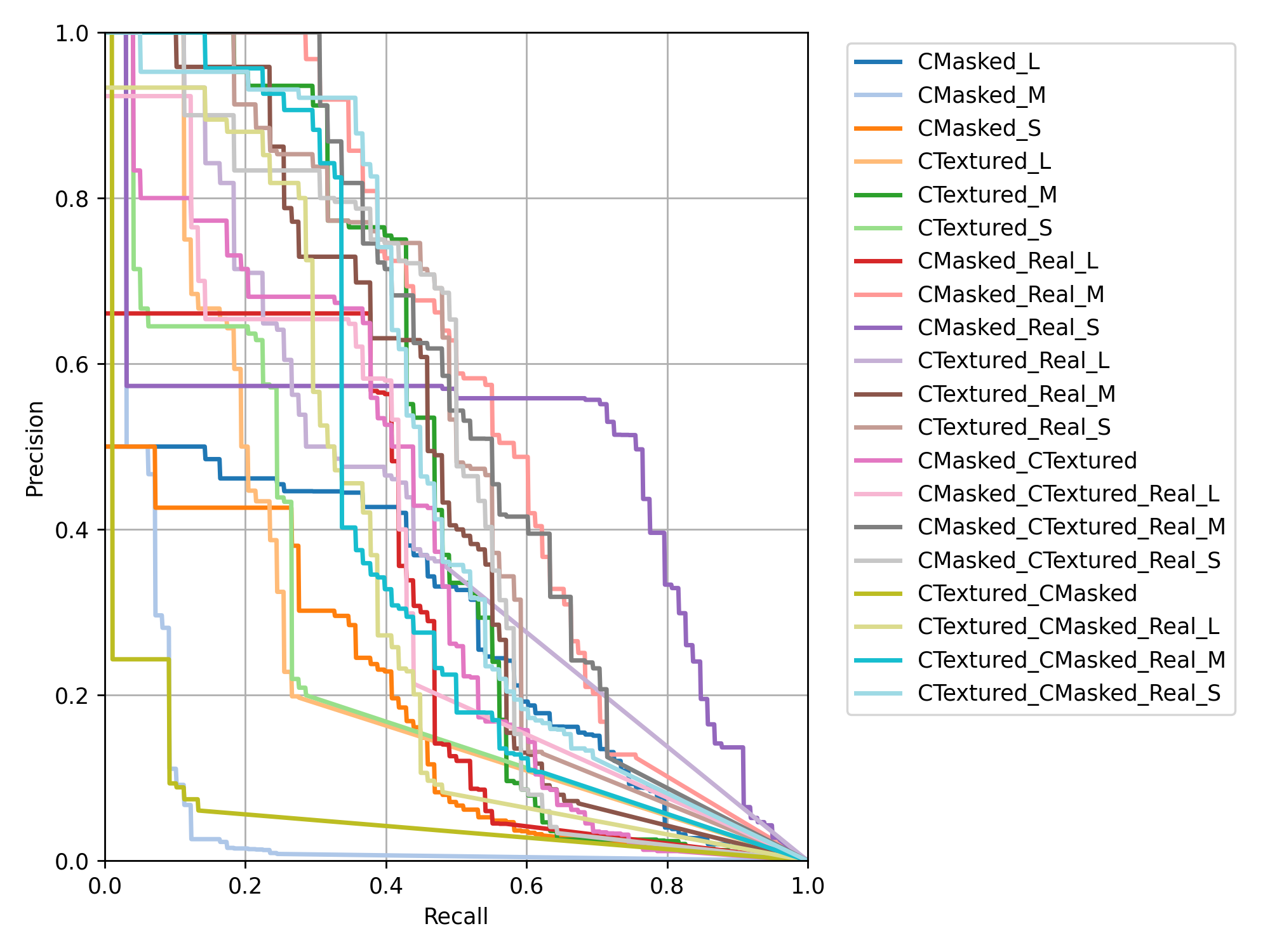}}
    \subfigure[UE5 Augmented]{\includegraphics[trim=0cm 0.4cm 0.5cm 0cm, clip, width=0.47\linewidth]{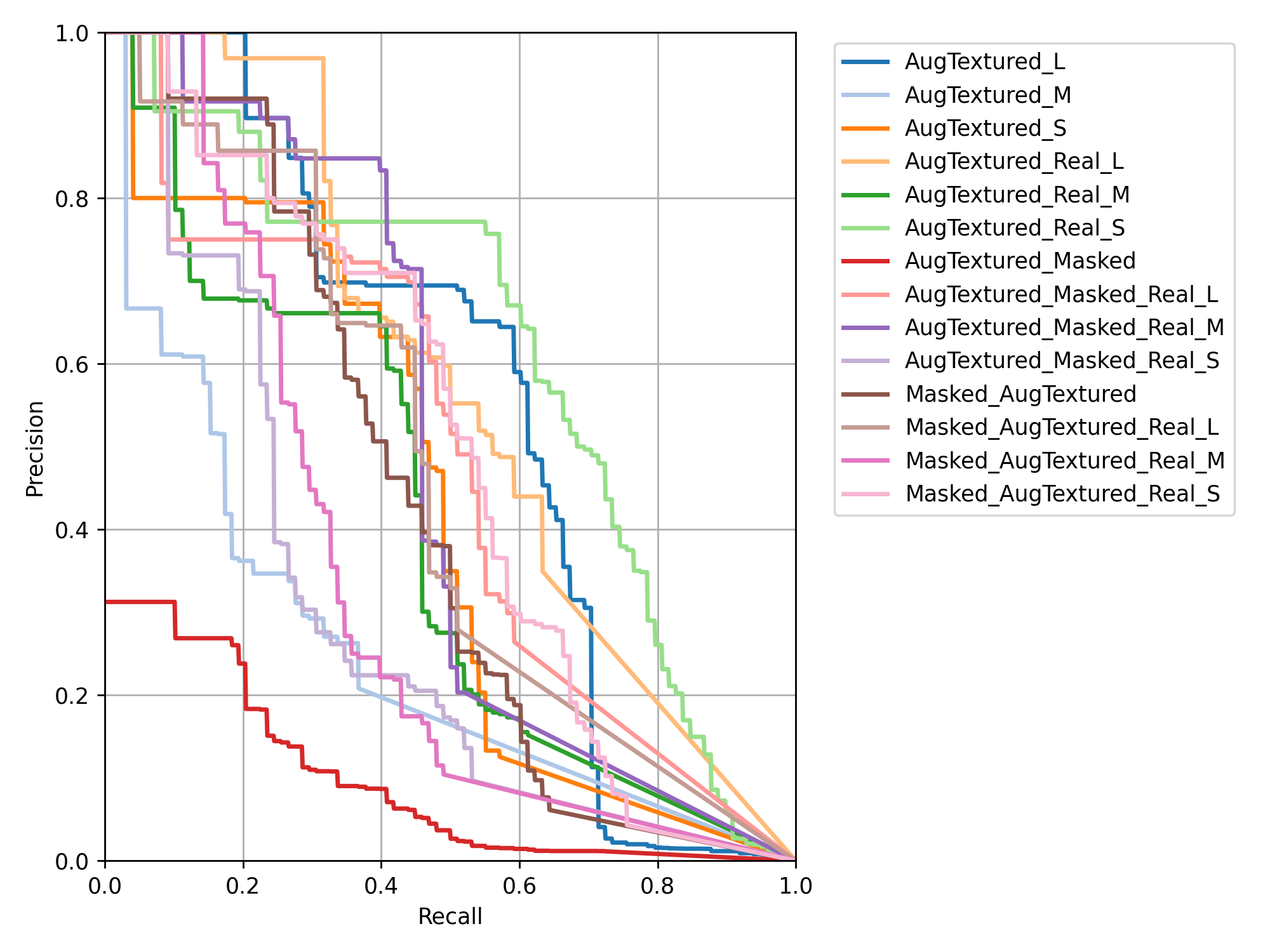}}
    \subfigure[UE4 \& UE5 Augmented]{\includegraphics[trim=0cm 0.4cm 0.5cm 0cm, clip, width=0.47\linewidth]{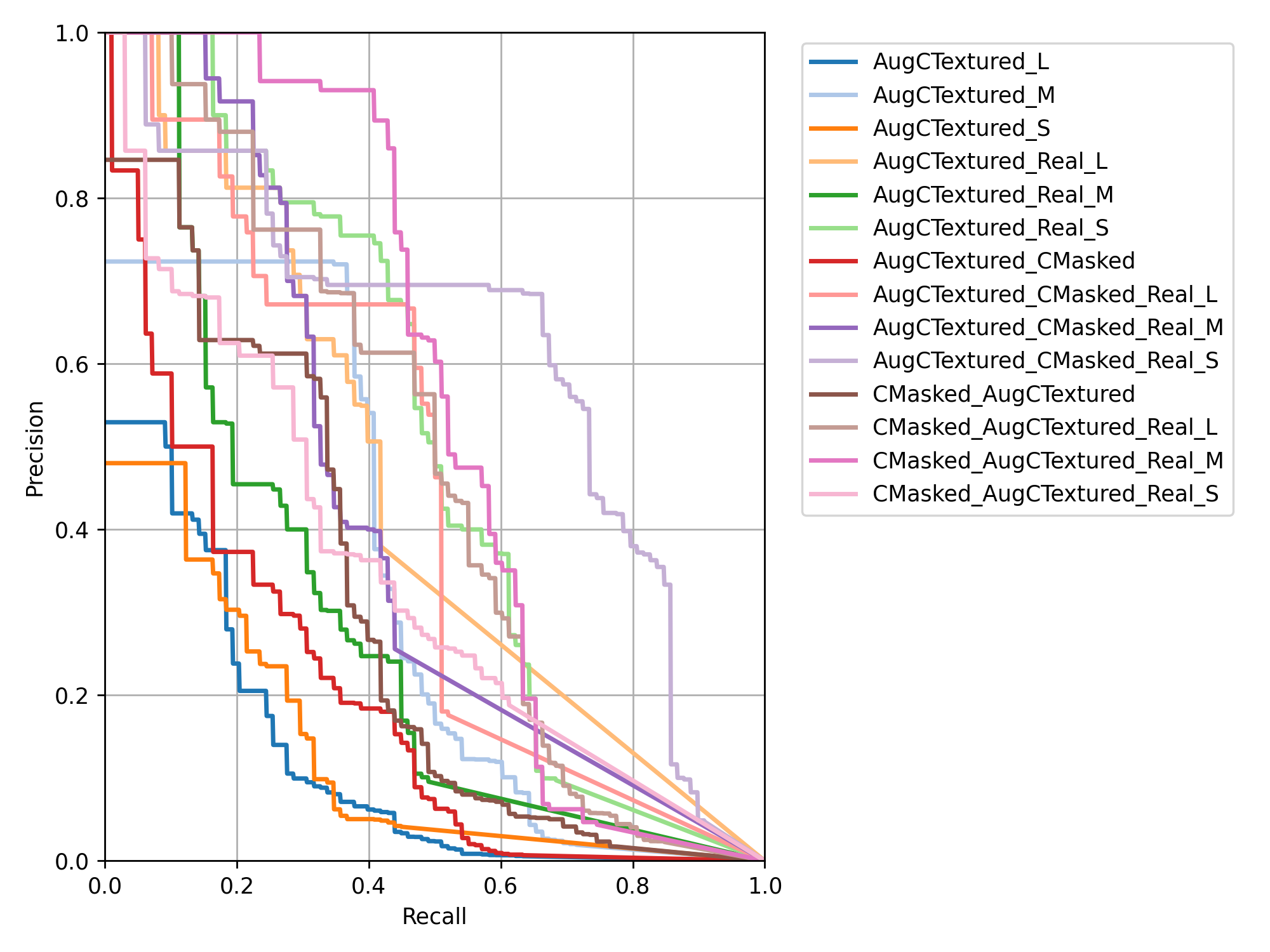}}
    \caption{The detection testing results (PR curves) of different YOLOv8n models trained on different data combinations for the shooter class.} 
    \label{fig:yolo_shooter}
\end{figure}

After obtaining the models by fine-tuning the pretrained YOLOv8n model on 71 different data combinations, we evaluate the performance by testing them on a dataset of 100 real images. We set the batch size to 1, the object confidence threshold for detection to 0.001, and IoU to 0.5 for the testing. The result of the shooter class detection is shown in Fig.~\ref{fig:yolo_shooter} and the result of combine shooter and gun detection is shown in Supplementary Fig. \hyperlink{sup:yolo-all}{S1}. We can see that by combining with Unreal Engine 4 (UE4) data and with the help of augmentation, the performance is improved compared to only using Unreal Engine 5 (UE5) data for training. Precision (P), recall (R), and mean average precision (mAP) results for the shooter and gun classes can be found in Table \ref{tab:shooter-all} and Supplementary Table \hyperlink{sup:gun-all}{S1}, respectively.

\begin{table}[ht!]
\centering
\caption{YOLOv8n testing results for the shooter class. Combination ID represented the order of training. The maximum amount of synthetic data was used for all sequential training scenarios, and S, M, and L amounts of real data correspond to 100, 300, and 500 images, respectively. For example, Textured\_Masked\_Real\_M is first trained on textured synthetic images, followed by masked synthetic images, and finished with 300 real images. The column labels represent the type of data used, signifying if UE5 or both UE4 and UE5 data is used and whether the textured synthetic data is augmented with camera sensor effects.}
\label{tab:shooter-all}
\adjustbox{width=\textwidth}{
\begin{tabular}{rcccccccccccc}
\hline
\multicolumn{1}{c}{\textbf{Combination}} &
  \multicolumn{3}{c}{\textbf{UE5}} &
  \multicolumn{3}{c}{\textbf{UE4 \& UE5}} &
  \multicolumn{3}{c}{\textbf{UE5 Augmented}} &
  \multicolumn{3}{c}{\textbf{UE4 \& UE5 Augmented}} \\ \hline
\multicolumn{1}{c}{\textbf{ID}} &
  \textbf{precision} &
  \textbf{recall} &
  \textbf{mAP50} &
  \textbf{precision} &
  \textbf{recall} &
  \textbf{mAP50} &
  \textbf{precision} &
  \textbf{recall} &
  \textbf{mAP50} &
  \textbf{precision} &
  \textbf{recall} &
  \textbf{mAP50} \\ \hline
\textbf{Masked\_L}                 & 0.125   & 0.13  & 0.0843   & 0.152  & 0.16   & 0.1    & -     & -      & -      & -      & -     & -      \\
\textbf{Masked\_M}                 & 0.00749 & 0.01  & 0.000763 & 0.056  & 0.02   & 0.0182 & -     & -      & -      & -      & -     & -      \\
\textbf{Masked\_S}                 & 0.0717  & 0.59  & 0.0792   & 0.071  & 0.09   & 0.0387 & -     & -      & -      & -      & -     & -      \\ \hline
\textbf{Textured\_L}               & 0.00625 & 0.84  & 0.019    & 0.202  & 0.19   & 0.0961 & 0.411 & 0.01   & 0.195  & 0.697  & 0.07  & 0.134  \\
\textbf{Textured\_M}               & 0.0831  & 0.124 & 0.0384   & 0.284  & 0.0198 & 0.103  & \textbf{0.743} & 0.0579 & 0.0754 & 0.389  & 0.19  & 0.16   \\
\textbf{Textured\_S}               & 0.00836 & \textbf{0.89}  & 0.11     & 0.0857 & 0.15   & 0.0508 & 0.364 & 0.14   & 0.143  & 0.246  & 0.22  & 0.161  \\ \hline
\textbf{Real\_L}                   & 0.732   & 0.273 & 0.401    & -      & -      & -      & -     & -      & -      & -      & -     & -      \\
\textbf{Real\_M}                   & 0.701   & 0.235 & 0.521    & -      & -      & -      & -     & -      & -      & -      & -     & -      \\
\textbf{Real\_S}                   & 0.00667 & 1     & 0.483    & -      & -      & -      & -     & -      & -      & -      & -     & -      \\ \hline
\textbf{Masked\_Real\_L}           & \textbf{0.785}   & 0.28  & 0.35     & 0.304  & 0.4    & 0.326  & -     & -      & -      & -      & -     & -      \\
\textbf{Masked\_Real\_M}           & 0.536   & 0.255 & 0.325    & 0.631  & 0.36   & 0.445  & -     & -      & -      & -      & -     & -      \\
\textbf{Masked\_Real\_S}           & 0.477   & 0.484 & 0.43     & 0.579  & \textbf{0.62}   & \textbf{0.579}  & -     & -      & -      & -      & -     & -      \\ \hline
\textbf{Textured\_Real\_L}         & 0.524   & 0.23  & 0.316    & \textbf{0.713}  & 0.273  & 0.391  & 0.714 & 0.249  & 0.331  & 0.687  & 0.28  & 0.336  \\
\textbf{Textured\_Real\_M}         & 0.771   & 0.201 & 0.312    & 0.533  & 0.31   & 0.34   & 0.509 & 0.331  & 0.323  & 0.784  & 0.27  & 0.43   \\
\textbf{Textured\_Real\_S}         & 0.557   & 0.315 & 0.311    & 0.474  & 0.23   & 0.24   & 0.468 & 0.42   & 0.405  & 0.522  & 0.26  & 0.271  \\ \hline
\textbf{Masked\_Textured}          & 0.0835  & 0.06  & 0.0507   & 0.134  & 0.06   & 0.0944 & 0.103 & 0.13   & 0.074  & 0.23   & 0.377 & 0.201  \\
\textbf{Masked\_Textured\_Real\_L} & 0.485   & 0.565 & 0.476    & 0.667  & 0.23   & 0.332  & 0.659 & 0.27   & 0.342  & 0.355  & \textbf{0.65}  & 0.517  \\
\textbf{Masked\_Textured\_Real\_M} & 0.664   & 0.27  & 0.361    & 0.595  & 0.35   & 0.395  & 0.69  & \textbf{0.378}  & \textbf{0.44}   & 0.555  & 0.41  & 0.421  \\
\textbf{Masked\_Textured\_Real\_S} & 0.501   & 0.43  & 0.379    & 0.486  & 0.26   & 0.283  & 0.391 & 0.36   & 0.332  & \textbf{0.803}  & 0.204 & 0.311  \\ \hline
\textbf{Textured\_Masked}          & 0.0945  & 0.06  & 0.0454   & 0.0533 & 0.09   & 0.034  & 0.174 & 0.19   & 0.0958 & 0.0398 & 0.09  & 0.0121 \\
\textbf{Textured\_Masked\_Real\_L} & 0.717   & 0.228 & 0.306    & 0.526  & 0.432  & 0.419  & 0.419 & 0.23   & 0.259  & 0.769  & 0.31  & 0.37   \\
\textbf{Textured\_Masked\_Real\_M} & 0.565   & 0.689 & \textbf{0.606}    & 0.543  & 0.36   & 0.361  & 0.636 & 0.24   & 0.314  & \textbf{0.803}  & 0.3   & 0.432  \\
\textbf{Textured\_Masked\_Real\_S} & 0.389   & 0.6   & 0.443    & 0.468  & 0.2    & 0.265  & 0.513 & 0.27   & 0.331  & 0.55   & 0.49  & \textbf{0.522}  \\ \hline
\end{tabular}
}
\end{table}

\subsection*{Tracking with Deep OC-SORT and OSNET ReID}

Evaluating tracking performance is much more tedious for custom data than detection performance. Rather than just frames, entire videos must be annotated frame by frame with consistent IDs throughout. Our preliminary tracking evaluation is limited to six real videos we annotated. Rather than only include videos with at least one shooter in it, we also included a video of a busy mall to challenge the false positive rate of the models. All 71 detection models were tried with individually varying confidence thresholds for the gun and shooter classes. Additionally, we ran each configuration for both tracking a shooter alone and tracking a shooter with our gun-based confirmation. The overall best-performing data combination combines UE4 and UE5 data with augmented textured synthetic data, shown in Fig. \ref{fig:shooter_tracking}. Results for the other data types can be seen in Supplementary Fig. \hyperlink{sup:shooter_tracking-all}{S2}.


\begin{figure}[ht!]
    \centering
    \subfigure[Shooter Only]{\includegraphics[trim=1.75cm 0.45cm 1.75cm 1.25cm, clip, width=0.47\linewidth]{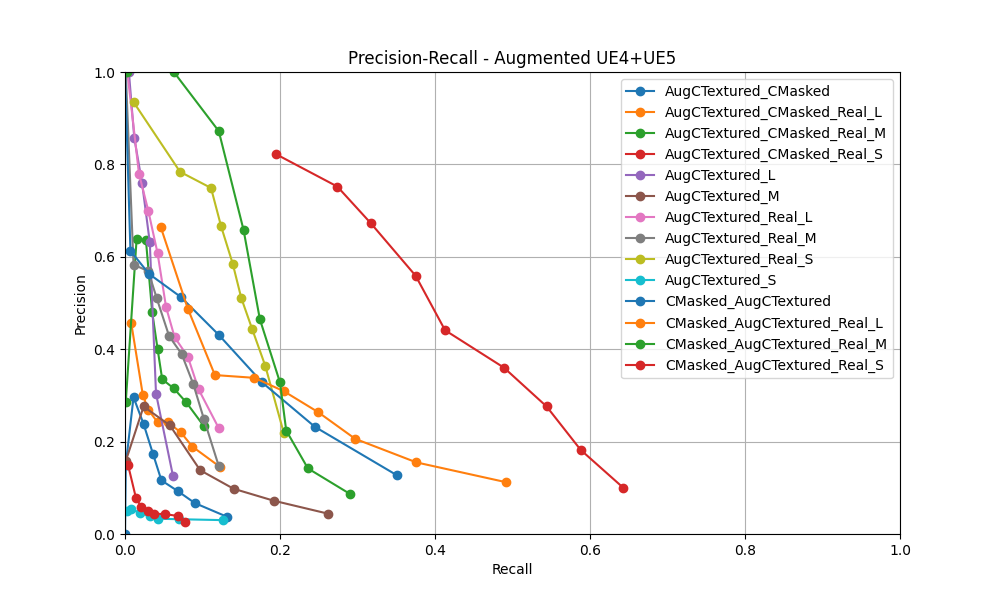}} 
    \subfigure[Shooter Confirmed with Gun]{\includegraphics[trim=1.75cm 0.45cm 1.75cm 1.25cm, clip, width=0.47\linewidth]{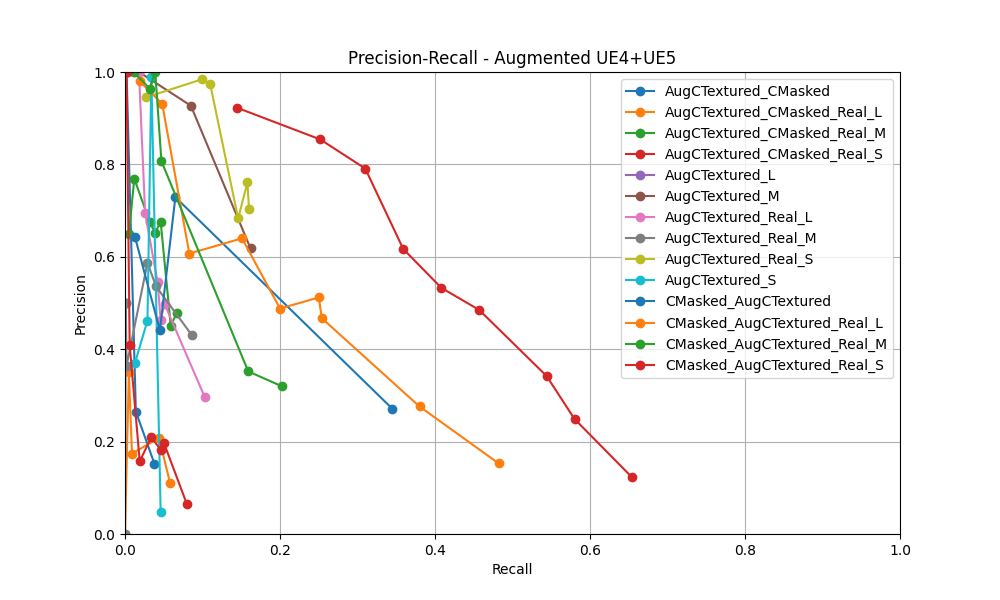}}
    \caption{The testing results (PR curves) for (a) tracking with only shooter detections and (b) tracking with shooter detections confirmed with a gun detection. Confidence thresholds for the gun and shooter detections were varied individually from 0.1 to 0.9. Only combined augmented models are shown here, the other data combinations are shown in Supplementary Fig. S2.} 
    \label{fig:shooter_tracking}
\end{figure}



The tracking results for the \textit{AugCTextured\_CMasked\_Real\_S}, with a gun confidence threshold of 0.8 and a shooter confidence threshold of 0.6, can be seen in Table~\ref{tab:tracking}. This table includes many of the standard multi-object tracking metrics, such as ID F1 score (IDf1), precision (IDP), and recall (IDR), regular recall (R) and precision (P), true positives (TP), false positives (FP), false negatives (FN), ID switches (IDSW), and multi-object tracking accuracy (MOTA). Metrics with an ID prefix correspond to the performance of maintaining the correct IDs. 

\begin{table}[ht!]
\centering
\caption{Tracking results using the YOLOv8n model, \textit{AugCTextured\_CMasked\_Real\_S}, trained with 2,545 augmented textured synthetic images from UE4 and UE5, followed by 12,698 masked synthetic images from UE4 and UE5, and finished with 100 real images. A gun confidence threshold of 0.8 and a shooter confidence threshold of 0.6 was used for these results.}
\resizebox{0.75\textwidth}{!}{%
\begin{tabular}{|c|c|c|c|c|c|c|c|c|c|c|}
\hline
\textbf{IDF1} & \textbf{IDP} & \textbf{IDR} & \textbf{R} & \textbf{P} & \textbf{TP} & \textbf{FP} & \textbf{FN} & \textbf{IDSW} & \textbf{MOTA} \\ \hline
0.302                          & 0.443                         & 0.229                         & 0.349                       & 0.674                       & 10                           & 432                          & 1,673                         & 24                             & 0.171                          \\ \hline
\end{tabular}%
\label{tab:tracking}
}
\end{table}

\subsection*{System-level performance}
Standard performance metrics alone do not comprehensively measure the system's performance in real-world use. To account for this, we also consider varying windows of frames around the ground truth where a bounding box would be considered. The intuition behind this is that while a constant track may not be achievable, consistent updates can provide valuable information. The number of frames in the window varies from 1 to 60, where the videos are 30 or 50 frames per second. 


\begin{figure}[ht!]
    \centering
    \includegraphics[width=0.75\linewidth]{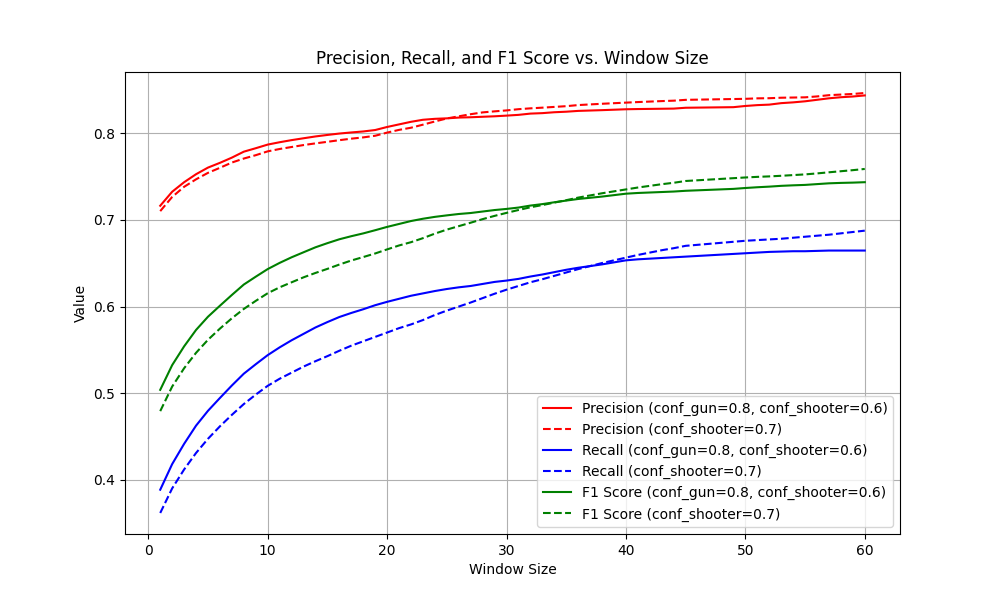}
    \caption{System-level performance (precision, recall, and F1 score) of the AugCTextured\_CMasked\_Real\_S model with a varying window of frames to consider for bounding box matches and with and without gun-based shooter confirmation.} 
    \label{fig:system_perf}
\end{figure}

\subsection*{Edge-device performance}
For this system to be useful, it must be able to run at high enough fps to capture quick movement through a sometimes relatively narrow field of view, such as running across a hallway. We measured the time required for each component's computation on a Raspberry Pi 4 (RPi4) with 4GB of RAM and Jetson Nano. The inferencing time for YOLOv8n and Deep OC-SORT are also tabulated in Table~\ref{tab:speed_results}. For both of these devices, a consumer $1920\times1080$ webcam was used as the video source, which was then padded and resized to $640\times640$ before being processed by YOLOv8 and Deep OC-SORT.

\begin{table}[ht!]
\centering
\caption{The inference speed results for running detection and tracking on the selected edge devices.}
\resizebox{0.6\textwidth}{!}{%
\begin{tabular}{|c|ccc|c|}
\hline
\multirow{2}{*}{\textbf{Device}} & \multicolumn{3}{c|}{\textbf{Computation Time per Image (ms)}}                                     & \multirow{2}{*}{\textbf{Total (FPS)}} \\ \cline{2-4}
                                 & \multicolumn{1}{c|}{\textbf{Detection}} & \multicolumn{1}{c|}{\textbf{Tracking}} & \textbf{Total} &                                       \\ \hline
\textbf{Raspberry Pi 4}          & \multicolumn{1}{c|}{650}                & \multicolumn{1}{c|}{250}               & 900              & 1.11                                     \\ \hline
\textbf{Jetson Nano}             & \multicolumn{1}{c|}{125}                & \multicolumn{1}{c|}{55}                & 180              & 5.56                                     \\ \hline
\end{tabular}%
}
\label{tab:speed_results}
\end{table}




 

\section*{Discussion}



After training 71 YOLOv8n models with the various data combinations, a few trends are noticeable. First of all, gun detection performance is inferior compared to shooter detection performance. There are a few potential causes, such as the task simply being more difficult than detecting shooters due to the smaller size of guns. It may also be due to not strict enough filtering of the synthetic data using a bounding box size threshold. Removing some of the very small instances of gun labels in the training set may allow the models to learn more effectively. Another approach would be to merge our data with an existing gun dataset. By merging the existing gun dataset in our model training, it provides greater data variability and potentially improves the detection performance and robustness of the model in detecting different types of guns. 
Another trend is that training with some synthetic followed by real data consistently performs better than training with only real data. For example, by comparing the performance of \textit{Real\_M} with \textit{Textured\_Masked\_Real\_M}, we can see that the mAP increased by 16.3\%.



It's difficult to determine which model is best for our application based only on the detection results, so we also tested all 71 models with varying confidence thresholds with tracking on six videos, one of which is of a busy mall with no shooters. We included the video of the mall to challenge the false positive rate of the models. Tracking also allows the use of our gun-based shooter confirmation system, which is discussed in more detail later. As expected, this system is very dependent on the quality of gun detection, which, while it does improve the performance of some models, it also collapses the performance of others. The overall best data combination for tracking is augmented UE4+UE5 and real data, with \textit{AugCTextured\_CMasked\_Real\_S} being the best model for both tracking with only the shooter and tracking with gun-based confirmation. Interestingly, a model trained with only 100 real samples, rather than 300 or 500, is the best performing. This may be due to the method of training where different data types are used sequentially, which could imply that the synthetic data better generalizes the models compared to training with a larger amount of the limited real data.
The tracking results for the other data combinations can also be seen in Fig.~\ref{fig:shooter_tracking}; however, these models seem particularly impacted by low gun detection performance. Regardless, some other trends are apparent. Training with UE5 and Real data performs relatively well, but adding UE4 data or augmenting UE5 data slightly decreases performance. However, adding \textit{augmented} UE4 data and \textit{augmented} UE5 improves performance. The obvious thing to try would be to add augmented UE4 data to unaugmented UE5 data. Lastly, the current system seems prone to ID switches and has difficulty maintaining a constant track. Regardless, tracking in its current form still reduces false positives, thus increasing robustness.








Another benefit of tracking rather than just detection is that it allows the further processing of a window of frames. This helps give a better sense of the real-world performance of the system where consistent, but not necessarily constant, updates can provide valuable information about a shooting event in real-time.

Higher FPS and resolution for a security system would always be ideal, but in practice, the additional cost of just the higher-quality cameras, not to mention the more powerful computing hardware that would be required, would make such a system hard to adopt. That being said, anecdotally, it seems that around 4 FPS or higher is sufficient to adequately capture the speeds at which a person can move through the view of a security camera, although more thorough testing would need to be done to verify this. Another consideration for tracking at lower FPS is approximations used to predict motion. The error associated with assuming constant velocity will increase as the time between frames increases.


\section*{Methods}

Our implementation of shooter tracking consists of three primary stages: (i) synthetic data generation, (ii) training YOLOv8, and (iii) tracking with Deep OC-SORT using OSNet ReID with gun detection-based shooter confirmation.

\subsection*{Synthetic data generation}
We generate synthetic data and perform domain randomization using Unreal Engine 4 and Unreal Engine 5~\cite{unrealengine} environments. The individual aspects of synthetic data generation will be discussed further in the subsections below.

\subsubsection*{Unreal Engine environment}
Unreal Engine 4 was used to simulate an active shooter's movements and those of evacuees, and the shooter was holding an assault rifle to differentiate them visually from the evacuees. The shooter did not fire their weapon during the simulation. Three sections of a hospital building were used, an open room, a hallway, and a staircase. Nodes were designated to correspond to specific locations inside the simulated environment, such as a junction point or an endpoint. Sample images from the environment can be seen in Fig.~\ref{fig:synth_data}. All actors, shooters, and evacuees were designed to move from one node to another. Evacuees were programmed to reach the nearest exit, and the shooter was programmed to reach a target node. The movement of the actors was facilitated through a navigation mesh, and the shooter was a dynamic obstacle on the mesh, so the evacuees tried to avoid the shooter. Cameras were placed strategically in the building to observe the shooter and evacuees. With proper camera placements, we could capture various movement interactions between the actors and the camera, such as the actors moving toward the camera, away from the camera, and perpendicular to the camera. 

The UnrealCV plugin~\cite{qiu2017unrealcv} was used to allow a Python script to modify the Unreal Engine 4 environment. This plugin enables us to modify the simulation settings, such as the position (location and rotation) and color of objects, along with the camera position and view type. There are two view types that we used in the simulation, the \textit{image} and \textit{object mask} view types, which represent the default textures and solid-colored segmentation masks, respectively. These capabilities allow us to randomize scenes with a shooter, evacuees, and multiple camera locations. On top of that, the images from both textured and masked images can be saved for further processing.

We also used Unreal Engine 5 to simulate an active shooter with civilians in various higher-fidelity environments. These environments consist of a school, a supermarket, and a bank; in which we recorded data from three locations in both the school and supermarket and four locations in the bank. The shooter can hold either a handgun or a rifle to increase the variety in the data. We also vary the number of civilians to have low- and high-density scenes. The civilian models are randomly generated from a pool of assets with adjustable parameters. Rather than limit the possible character positions through creating a realistic shooting simulation, we choose to have the civilians move in essentially random paths along with randomly placing the shooter. This creates more challenging scenarios where the shooter is partially occluded. However, this data is only useful for training detection since there is no continuity between frames. Sample images from the environments can be seen in Fig.~\ref{fig:synth_data}.

\subsubsection*{Domain randomization}
While synthetic data is an amazing tool for deep learning, some care needs to be taken when using it to train models. There are domain differences between real and synthetic data; thus, when synthetic data is used for training, it generally cannot be expected to work directly for inference on real data. Closing this gap between domains is called domain adaptation, for which many different techniques exist. Probably the most intuitive technique is to have a high-fidelity simulation to appear as realistic as possible. While we try to have fairly high-fidelity synthetic data, especially in the case of the UE5 environments, we also choose to use domain randomization due to the level of control achievable with the Unreal Engine environments. This approach allows for simple yet effective implementation where we randomly sample positions for the shooter and evacuees within bounded areas and randomly sample colors for everything in the environment.





The synthetic data generation process with Unreal Engine 4 can be seen in Fig.~\ref{fig:synth_data}. The environment initializes with the actual textures of the environment. The first step is to update the position of the actors, which includes the evacuees and the shooter. We achieved this by randomly sampling positions within the bounded area (shown in green). The camera positions and viewing angles are randomized within smaller valid regions (shown in red). From here, textured synthetic (TS) images are exported before randomizing the colors of everything in the environment. The colors are changed by switching to the object mask view and randomly setting the red, green, and blue (RGB) channel values to between 0 and 255 for each object using UnrealCV. The resulting scene is exported as domain-randomized masked synthetic (MS) images. Then, the selected colors for the shooter and guns are used to easily threshold the segmented images to generate tight bounding box annotations. We found this extra step to manually create precise bounding boxes necessary because bounding boxes created within Unreal Engine 4 would often have a hand or foot of the shooter partially outside. Finally, the view is switched back to default with actual textures, and the randomization loop restarts from the beginning. 

This configuration allows us to easily make adjustments to the overall process. One such adjustment is that instead of randomizing the position of the actors in the UE4 TS data, we freeze and unfreeze the simulation to capture time-continuous data. Additionally, the TS output images don't necessarily need to be saved when generating MS data. However, the MS images are still temporarily required to generate the bounding box annotations for the TS images.

The generation process with Unreal Engine 5 can be seen in Fig.~\ref{fig:synth_data} and consists of running a short simulation in each desired area. The cameras are positioned to emulate the view of actual security cameras. However, to increase variety in the data, we slightly perturb the position and viewing angle for each image. Textured and masked data are captured separately, and bounding box annotations are generated directly using Unreal Engine 5. Instead of varying the colors of everything in Unreal Engine, we keep constant mask colors throughout the simulation and randomize the colors in a separate Python script using simple thresholding based on the already masked images. This significantly speeds up the data generation process because Unreal Engine would need to cycle through every object in the environment rather than just the visible masks.

\subsubsection*{Camera sensor effects}
\begin{figure}[h!]  
  \centering
  \includegraphics[width=1\linewidth]{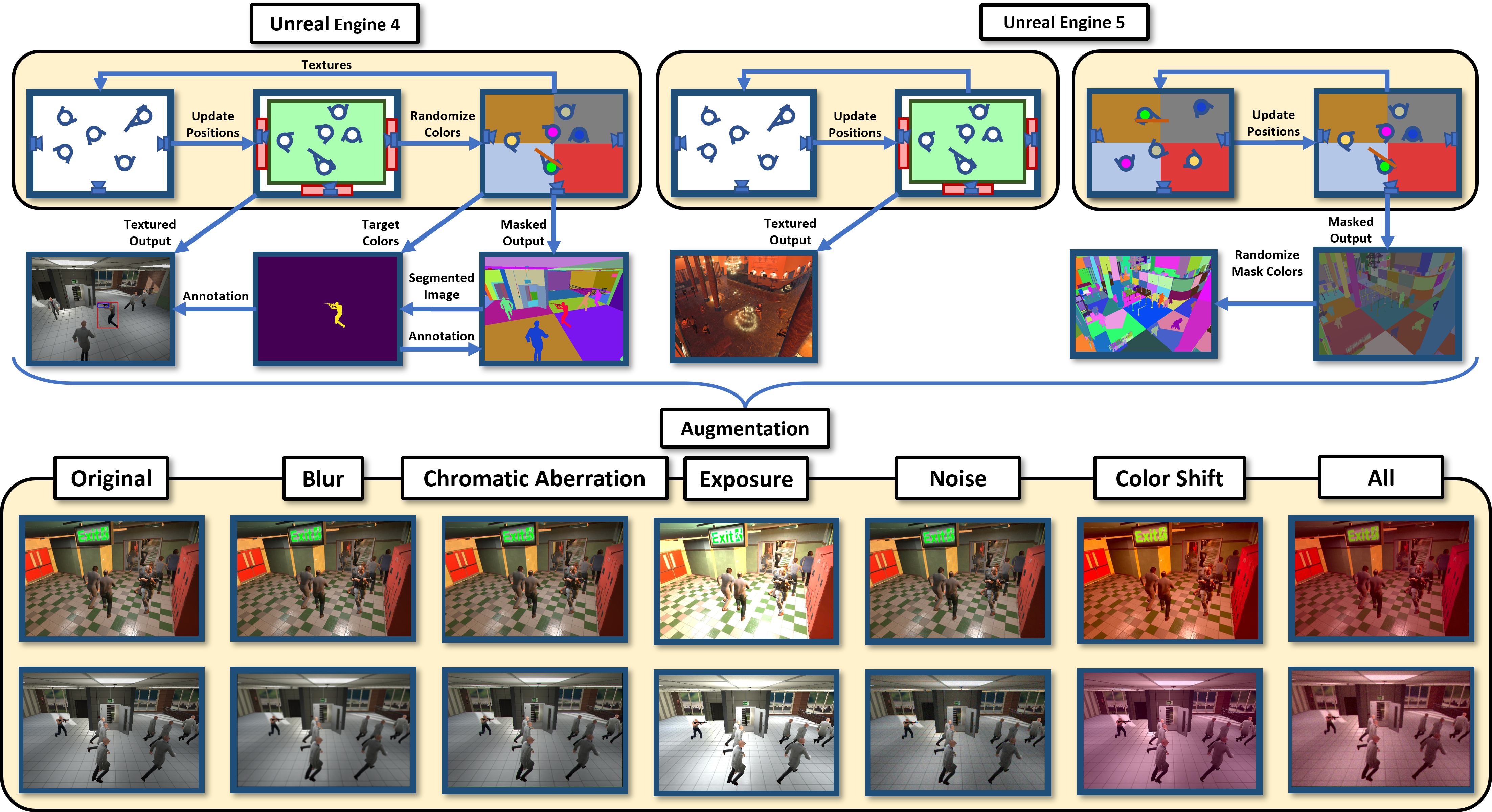}
  \caption{\label{fig:synth_data}
  Complete synthetic data generation process. We utilize the UnrealCV plugin for UE4 to generate textured and masked synthetic data to obtain accurate bounding boxes by thresholding the masked images. Textured and masked data are generated separately in UE5 as we are able to extract accurate bounding boxes directly. Both UE4 and UE5 textured images are augmented with camera sensor effects.}
\end{figure}

We augment the textured synthetic data using camera sensor effect modeling~\cite{carlson2018modeling}. Rather than applying all the effects on all the images uniformly, they are applied at random levels. The objective of augmenting the textured synthetic data is to help blend the differences between synthetic and real data by making it more similar to real data. We chose to augment only textured data rather than the masked data to limit the already large number of combinations. It's possible that augmenting the masked data would further improve its ability to transfer to real data. This is done by breaking up aspects of synthetic data, such as extremely well-defined edges. Adding noise mimics the normal noise found in real images introduced by limitations in the camera's sensor. Blurring the images helps make the edge less defined. Chromatic aberration mimics an effect along the edges of objects caused by camera lenses. Adjusting exposure helps account for varying camera quality and lighting changes throughout the day. Lastly, color shift helps account for different camera sensors, where one may be more sensitive to certain colors than another. These camera sensor effects can be seen in Fig.~\ref{fig:synth_data}.



\subsection*{Training YOLOv8}
YOLOv8~\cite{Jocher_YOLO_by_Ultralytics_2023} is one of the latest versions of the popular you only look once (YOLO) object detection framework. It has made several advancements upon previous versions that significantly improve its performance with small models, making it well-suited for edge devices.

The classes we aim to detect are shooter and gun, where the shooter class bounding boxes contain both the person and the visible portion of the gun. If the gun becomes completely obscured, we still label the person as a shooter based on prior knowledge and subtle posture hints, such as hunched shoulders. While we are primarily concerned with tracking the shooter, including gun detection as confirmation of a new shooter may help reduce false positives. We have two groups of image data: real images extracted from videos publicly available online and synthetic images created using the Unreal Engine environments. The synthetic data can be further divided into two subgroups: semi-realistic textured synthetic images (TS) and masked synthetic (MS) images. Both sets of synthetic data are created from scenes where the characters' positions are randomized for each frame, with animations updating their appearance as if they were moving. 

Our dataset includes 700 real images, split into 500 training, 100 validation, and 100 testing images. Synthetic data with semi-realistic textures consists of 1,567 images from UE5 and 978 images from UE4, and synthetic data with masked textures includes 7,415 images from UE5 and 5,283 images from UE4. All synthetic data is used only for training. We also explore the use of camera sensor effect augmentation for the TS data. All synthetic data is used only for training. The object detection ground truth contains the $x$ and $y$ coordinates of the center of the bounding box localizing the shooter or gun, as well as the width and height of the bounding box. The training image size is the default value of $640\times640$, so the images are resized and padded before training.

Although not exhaustive, we conducted a series of training experiments to find the best combination of these three image data types: real (R), textured synthetic (TS), and masked synthetic (MS). For each combination, different numbers of images are used. We train four sets of 23 different data combinations, as seen in the combination column of Table~\ref{tab:shooter-all}. The four sets include data generated with UE5, data generated with UE4 and UE5, augmented data generated with UE5, and augmented data generated with UE4 and UE5. Rather than starting training from scratch, we use the YOLOv8n weights pre-trained on the COCO dataset~\cite{lin2015microsoft}. These weights can be used to detect 80 classes of objects found in the COCO dataset. We run training for the default 100 epochs with early stopping and patience value of 
50 epochs.
When multiple data types are being used, the first type of data starts with the pre-trained weights and trains for 100 epochs, then the following type of data resumes the training with the weights obtained from the previous training phase and trains for another 100 epochs.


\subsection*{Tracking with Deep OC-SORT, OSNET ReID, and shooter confirmation with gun detections}
The YOLO tracking toolbox~\cite{Brostrom_Real-time_multi-object_tracking} was immensely helpful when exploring the performance of different tracking algorithms on a Jetson Nano. We chose to use Deep OC-SORT \cite{maggiolino2023deep} with OSNET \cite{zhou2019omniscale} using x0 25 MSMT17 weights for the re-identification model for the balance of speed and accuracy, even at a low framerate. As its name suggests, Deep OC-SORT builds upon OC-SORT \cite{cao2023observationcentric}, which addresses some limitations of SORT \cite{Bewley_2016}. One of the limitations of SORT is that it is a purely motion-based tracker. This means that when an object is lost, the estimated location from the Kalman filter will likely deviate from the actual location as time continues. This means that even when the object is detected again, it will not be part of the same track. Observation-centric Re-Update (ORU) reduces the accumulated error by backchecking and updating the parameters of the Kalman filter when an object is detected again. This re-update is based on virtual trajectories of the untracked period. Observation-Centric Momentum (OCM) aims to consider the size of $\Delta$t used to estimate the velocity. While a small $\Delta$t is necessary for the linear-motion assumption, a large $\Delta$t also increases noise in the measurement; thus, the choice to increase it comes at a trade-off. Lastly, Observation-Centric Recovery (OCR) starts a second association attempt between the last unmatched observations and tracks. This can handle objects stopping or being occluded for a short duration. 
Deep OC-SORT improves upon OC-SORT in a few ways. Firstly, they implement camera motion compensation to adjust predicted bounding boxes based on the camera's movement. However, we disable camera motion compensation to slightly improve inference speed because the cameras in our application are stationary. Secondly, their implementation of appearance association is dynamic and linearly scales based on the confidence of the detection. Lastly, adaptive weighting is used to increase the weight of appearance features depending on the discriminativeness of the embeddings. This is used to boost track-box scores for cases where there is a high degree of similarity.

\begin{figure}[h!]  
  \centering
  \includegraphics[width=1\linewidth]{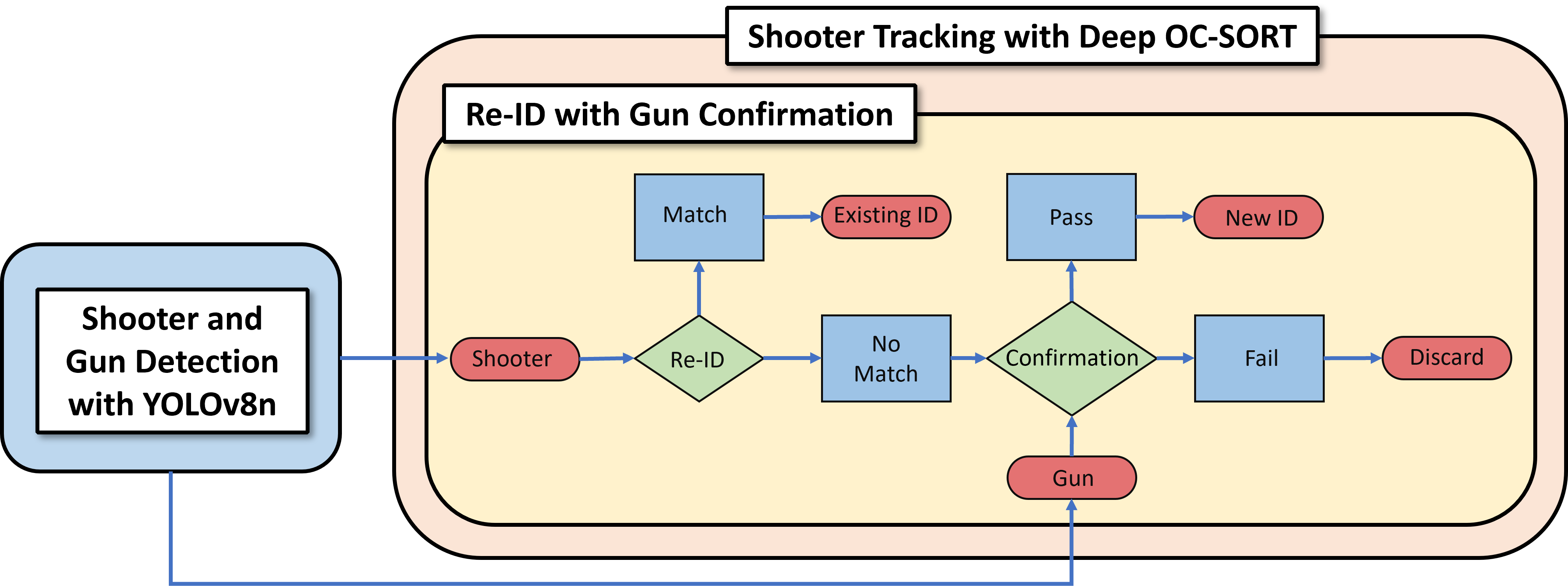}
  \caption{\label{fig:gun_confirm}
  The pipeline of our gun confirmation system for shooter tracking. Shooter and gun detection boxes are sent to the Deep OC-SORT association phase. The gun detections are not tracked but are instead used in track initialization to confirm a new shooter detection before assigning that track an ID. After a shooter has an ID, that track no longer requires a gun detection to continue being tracked. If a shooter detection does not match an existing ID and does not have a gun detection to confirm it, it is discarded.}
\end{figure}

While detecting and tracking shooters as a whole allows for more robust tracking when compared to only guns, it also increases the likelihood of falsely detecting a regular person as a shooter. To address this, we require that a gun detection overlaps with a shooter detection before we begin tracking that person as a shooter. We are able to do this because our shooter class contains both the person and the gun when it is visible. After this confirmation step, we no longer require gun detection to continue tracking the shooter as long as the ReID model associates them with the same ID. If a new potential shooter ID is introduced, a gun detection will be required before that ID is labeled as a shooter. A new ID doesn't necessarily mean a new shooter since the ReID model can be confused by cases where the appearance of a shooter changes significantly between frames. This system is implemented for the track initialization stage of Deep OC-SORT; as such, a shooter ID can only be introduced in that stage. However, once the ID exists, it can be used in the first and second association stages of Deep OC-SORT to keep track of the shooter through occlusions or other situations where the detection accuracy decreases. This process can be seen in Fig.~\ref{fig:gun_confirm}.

\section*{Availability of materials and data}
The datasets used and/or analyzed during the current study are available from the corresponding author upon reasonable request.


\bibliography{references}



\section*{Acknowledgements}

This work was supported by the National Science Foundation Award\# CNS-1932033.






\section*{Author contributions statement}
Conceptualization, J.R.W., S.Y.T., and S.S.; 
Methodology, J.R.W.; 
Software, J.R.W.; 
Validation, J.R.W. and J.F.; 
Investigation, J.R.W. and J.F.; 
Data Curation, J.R.W., R.T. and L.H.; 
Writing - Original Draft, J.R.W. and J.F.; 
Writing - Review \& Editing, all authors; 
Visualization, J.R.W. and J.F.; 
Supervision, S.C. and S.S.; 
Project Administration, S.C. and S.S.; 
Funding Acquisition, S.C. and S.S.

\section*{Additional information}
\subsection*{Competing interests}
The authors declare no competing interests.




\includepdf[pages=-]{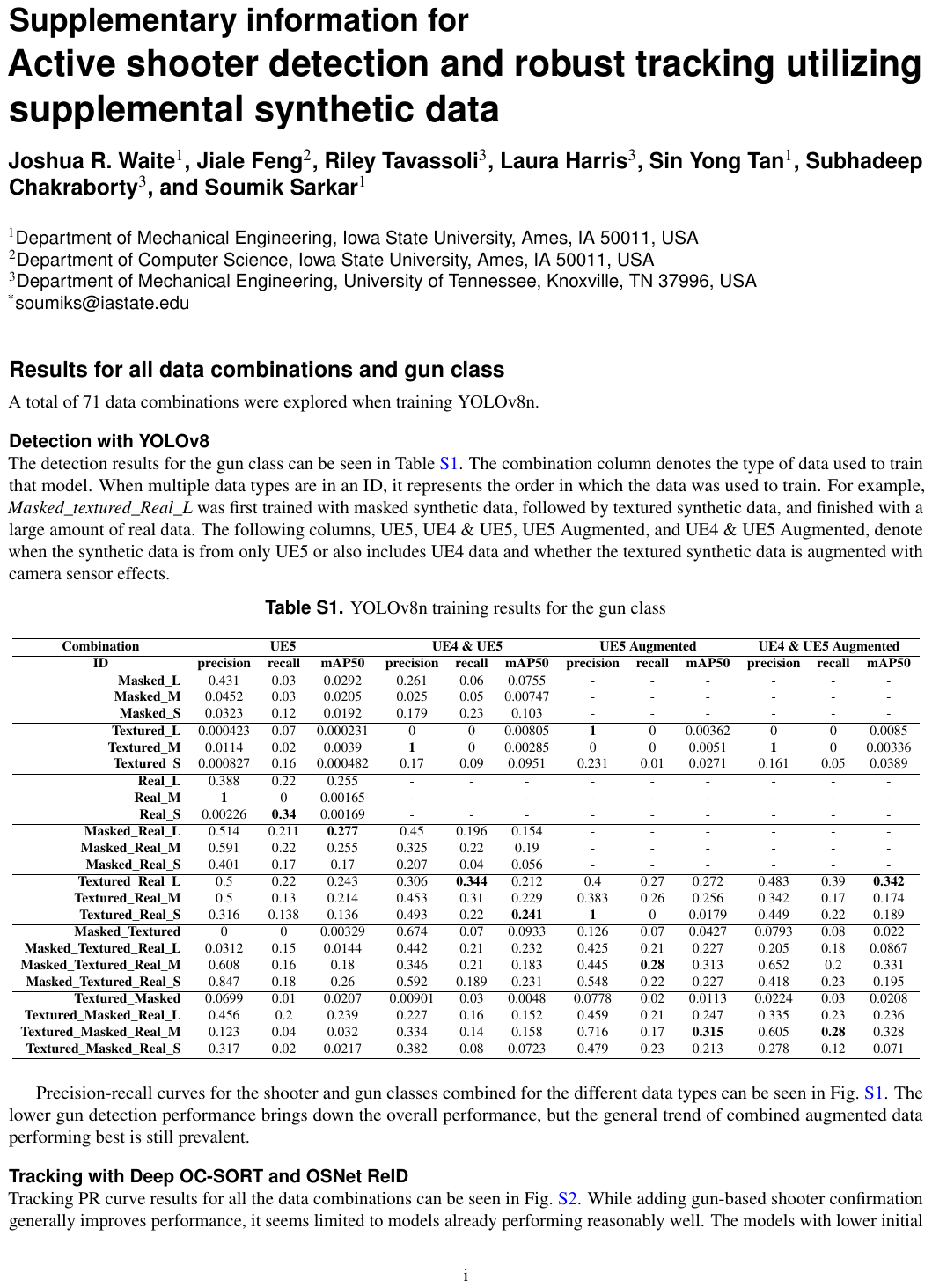}

\setcounter{page}{\value{page}-3}

\end{document}